# LoRA-SP: Streamlined Partial Parameter Adaptation for Resource-Efficient Fine-Tuning of Large Language Models

Yichao Wu*[a], Yafei Xiang[a], Shuning Huo[b], Yulu Gong[c], Penghao Liang[a]
[a]Northeastern University, 360 Huntington Ave, Boston, MA, USA 02115-5000;
[b]Virginia Tech, 600 Drillfield Dr, Blacksburg, VA, USA 24061-0001;
[c]Northern Arizona University, S San Francisco St, Flagstaff, AZ, USA 86011-0002
* Corresponding author: wu.yicha@northeastern.edu

## ABSTRACT

In addressing the computational and memory demands of fine-tuning LargeLanguage Models (LLMs), we propose LoRA-SP (Streamlined Partial Parameter Adaptation), a novel approach utilizing randomized half-selective parameter freezing within the Low-Rank Adaptation (LoRA) framework. This method efficiently balances pre-trained knowledge retention and adaptability for task-specific optimizations. Through a randomized mechanism, LoRA-SP determines which parameters to update or freeze, significantly reducing computational and memory requirements without compromising model performance. We evaluated LoRA-SP across several benchmark NLP tasks, demonstrating its ability to achieve competitive performance with substantially lower resource consumption compared totraditional full-parameter fine-tuning and other parameter-efficient techniques. LoRA-SP's innovative approach not only facilitates the deployment of advanced NLP models in resource-limited settings but also opens new research avenues intoeffective and efficient model adaptation strategies.

**Keywords:** Fine-tuning, Large Language Model, NLP, LoRA

## 1. INTRODUCTION

The advent of Large Language Models (LLMs) has revolutionized the field of natural language processing (NLP), propelling advances across a spectrum of tasks from natural language understanding (NLU) to natural language generation (NLG) (Brown et al., 2020[1]; Touvron et al., 2023[2]). The efficacy of these models, particularly when fine-tuned for specific downstream applications, has been well-documented, underscoring their adaptability and alignment with nuanced human intents (Liu et al., 2019[12]; Ouyang et al., 2022[14]). However, the process of fine-tuning, which traditionally involves the adjustment of all model parameters, imposes prohibitive computational and memory demands, especially as the scale of LLMs continues to grow. For instance, fine-tuning a model like LLaMA-65B with contemporary optimization methods requires over 1TB of GPU memory, a requirement that is far beyond the reach of most research and development entities[3].

To circumvent the exorbitant costs associated with full-parameter fine-tuning, the NLP community has turned towards Parameter-Efficient Fine-Tuning (PEFT) techniques. These methods, such as adapter weights and prompt weights adaptation, have demonstrated considerable promise in updating only a small fraction of a model's parameters, thus reducing the overall computational footprint (Houlsby et al., 2019[7]; Hu et al., 2022[8]; Li, 2021[11]). Among these innovations, Low-Rank Adaptation (LoRA) emerges as a particularly compelling solution, effectively balancing performance with efficiency by introducing low-rank matrices that approximate the changes to a model's weights during the fine-tuning phase (Hu et al., 2022[8]).

Despite its advantages, LoRA is not without its limitations, most notably its still significant activation memory consumption. This drawback stems from the necessity to store large input activations throughout the feed-forward and back-propagation phases, undermining its efficiency and limiting its applicability in resource-constrained settings[13]. Recognizing these challenges, we propose a novel adaptation of the LoRA methodology, LoRA-SP (Streamlined Partial Parameter Adaptation), which further optimizes the fine-tuning process by employing a half-selective parameter freezing strategy. This approach is inspired by the principles underlying dropout techniques in neural networks, which enhance generalization by randomly omitting units during training. Similarly, LoRA-SP selectively freezes half of the parameters during the fine-tuning process, inherently reducing memory demands and mitigating overfitting[23]. This strategic reduction not only optimizes computational resources but also exploits the inherent redundancy in LLMs, allowing for a more nuanced and efficient parameter adjustment. By integrating LoRA-SP into the fine-tuning pipeline, we aim to set a new benchmark for PEFT methods, combining enhanced efficiency with robust model performance.

LoRA-SP builds on the hypothesis that not all parameters need to be adapted for a model to learn task-specific nuances effectively. By randomly selecting half of the parameters for freezing, LoRA-SP leverages the intrinsic redundancy within LLMs, allowing for a more resource-efficient adaptation process. This method not only addresses the computational and memory inefficiencies inherent in existing PEFT methods but also introduces a scalable framework for fine-tuning LLMs across a diverse array of tasks and languages[4].

To validate the efficacy of LoRA-SP, we embarked on an extensive experimental campaign, fine-tuning models like RoBERTa, T5, and LLaMA across various benchmarks, including natural language understanding tasks, machine translation, and more specialized applications like long-context modeling. Our findings reveal that LoRA-SP achieves comparable, if not superior, performance to both full-parameter fine-tuning and traditional LoRA, all the while significantly reducing the memory footprint and computational overhead. Furthermore, our analysis delves into the hyperparameter sensitivity of LoRA-SP, showcasing its robustness and adaptability across different model architectures and tasks[18].

In summary, by selectively freezing half of the parameters, LoRA-SP significantly reduces both trainable parameters and activation memory requirements without compromising model performance. This method not only addresses the limitations of existing Parameter-Efficient Fine-Tuning (PEFT) methods but also offers a scalable, efficient framework for fine-tuning across various tasks and languages. Through this work, we aim to democratize access to state-of-the-art NLP technologies, facilitating their adoption across a wider range of applications and industries[10].

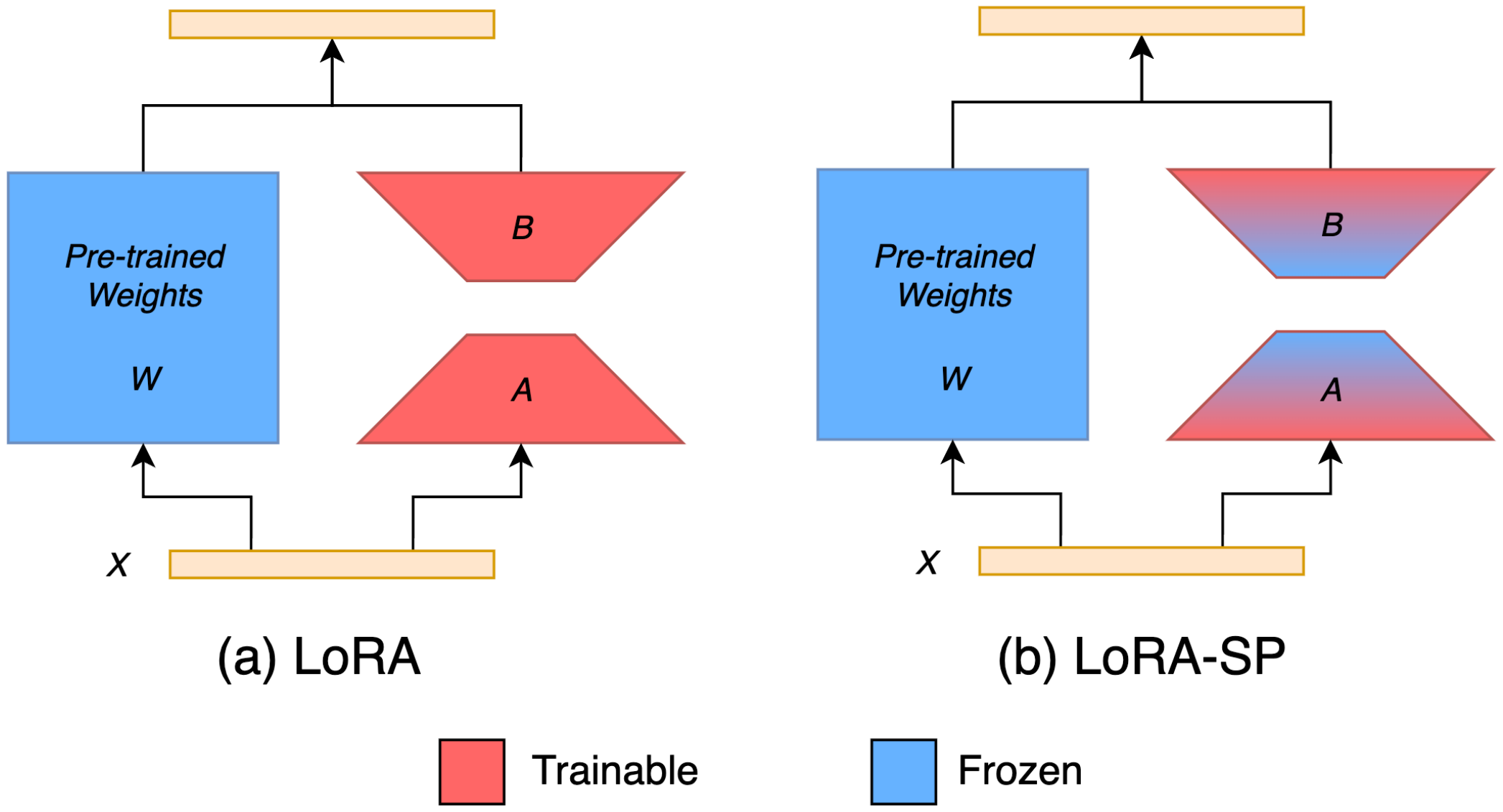

Figure 1. LoRA Framework. Left: Traditional LoRA. Right: LoRA-SP, partial parameter selected.

## 2. RELATED WORKS

### 2.1 Large Language Models (LLMs) and Transformer Architectures

The introduction of transformer-based LLMs has significantly advanced the field of natural language processing (NLP), showcasing remarkable capabilities in a wide range of tasks from machine translation to language modeling[1]. These models, exemplified by GPT and its successors, rely on deep stacks of transformer blocks comprising multi-head attention (MHA) and feed-forward network (FFN) modules. These architectural innovations have paved the way for models that understand and generate human-like text, leveraging vast amounts of parameters to capture intricate language patterns[3].

### 2.2 Low-Rank Adaptation (LoRA)

As the scale of LLMs grows, so does the challenge of fine-tuning them for specific tasks without incurring prohibitive computational and memory costs. The Low-Rank Adaptation (LoRA) technique emerges as a solution to this challenge by introducing low-rank matrices to model updates during fine-tuning, significantly reducing the number of parameters that need to be adjusted (Hu et al., 2022[8]). LoRA's strategy of adapting only a fraction of model parameters has demonstrated effectiveness across various models and tasks, striking a balance between fine-tuning efficiency and model performance. However, despite its advantages, LoRA faces limitations in managing activation memory consumption, a critical factor in computational efficiency and scalability (Zhang et al., 2023[21]).

### 2.3 Quantization and Efficiency in Fine-Tuning

Quantization techniques have been proposed to compress the model weights of LLMs, thereby reducing storage and memory requirements during both training and inference phases. These methods, including the NormalFloat Quantization[2], aim to maintain model performance while significantly lowering the resource footprint of LLMs. The integration of quantization with parameter-efficient fine-tuning methods like LoRA poses an interesting area of research, potentially offering a path to more resource-efficient adaptation of quantized LLMs for downstream tasks.

## 3. LORA-SP METHODOLOGY

### 3.1 Introduction to LoRA-SP

LoRA-SP, standing for Partial-Selective Low-Rank Adaptation, extends the conventional Low-Rank Adaptation (LoRA) approach by introducing a strategic partial freezing mechanism during the fine-tuning of large language models (LLMs). This mechanism aims at enhancing computational efficiency, reducing memory usage, and preserving, if not improving, the fine-tuning performance of LLMs[15]. The innovation lies in its selective approach towards parameter adaptation, choosing to update only half of the parameters within the low-rank matrices A and B, while the remainder is kept frozen. This section delves into the methodology behind LoRA-SP, detailing its unique adaptation strategy, integration with memory optimization techniques, and the implications on gradient compression[17].

### 3.2 Theoretical Justification and Derivation

Drawing parallels with the dropout technique in neural networks, which randomly omits units from the network during training to prevent overfitting, LoRA-SP selectively freezes parameters to achieve a similar regularization effect. This strategy is mathematically justified by considering the model's generalization ability as a function of its parameter variability. By freezing a subset of parameters, LoRA-SP introduces variability in the training process, enhancing the model's ability to generalize and reducing the likelihood of overfitting.

At the heart of LoRA-SP is the selective freezing of parameters, a novel strategy designed to optimize the fine-tuning process. The method employs a binary selection matrix S, which serves as a guide for identifying which parameters in the matrices A and B are to be adapted. Specifically, LoRA-SP targets half of the parameters for adaptation, thus significantly reducing the computational load and memory requirements. The adaptation process is formalized as follows:

$$S_{ij} = \begin{cases} 1, & if\ parameter\ ij\ is\ selected\ for\ adaptation, \\ & 0, otherwise. \end{cases} \tag{1}$$

To ensure a balanced selection, parameters for adaptation are chosen such that exactly half of the parameters in A and B are updated, while the other half remains frozen. This is achieved through a systematic selection process, ensuring that the trainable and non-trainable parameters are evenly distributed across the matrices. The adapted weight matrix is $\Delta W$ computed as:

$$\Delta W = (A \odot S)(B \odot S)^T. \tag{2}$$

where $\odot$ denotes the element-wise multiplication, thereby allowing only the selected half of the parameters to contribute to the model's adaptation.

### 3.3 Optimization of Memory and Computational Resources

To further enhance the efficiency of LoRA-SP, the methodology incorporates advanced memory optimization techniques, including weight quantization and selective activation recomputation:

**Quantization:** Applied to the non-trainable weights, quantization reduces the memory footprint by converting these weights into a compressed, quantized format. The operation is defined as:

$$Q = qN(W - (A \odot S)(B \odot S))^T. \quad (3)$$

where qN represents the quantization function, optimizing storage without affecting the fine-tuning performance significantly.

**Selective Activation Recomputation:** To manage memory usage effectively, LoRA-SP employs selective activation recomputation during the backward pass. This technique selectively recomputes only the necessary activations, avoiding the storage of all intermediate activations from the forward pass, thereby optimizing memory utilization.

### 3.4 Algorithmic Implementation of LoRA-SP

The implementation of LoRA-SP is articulated through a series of steps designed to ensure efficient fine-tuning:

Algorithm: LoRA-SP Fine-Tuning Process

Input: Pre-trained model weights W, binary selection matrix S for half parameter selection, target rank r, and the quantization function qN.

Initialization:

- Initialize low-rank matrices A and B, applying S to select half of the parameters for training.
- Quantize the non-trainable weights to reduce memory usage: Q = qN(W).

Fine-Tuning:

- For each training iteration:
  - Update selected parameters in A and B using gradient descent, as indicated by S.
  - Apply forward propagation with adapted weights $(W - (A \odot S)(B \odot S))^T$.
  - Utilize selective activation recomputation to manage memory efficiently during the backward pass.

This enhanced methodological description encapsulates the essence of LoRA-SP, emphasizing its selective adaptation mechanism that targets half of the parameters for updating. It elucidates the methodology's logical structure, from the initialization of selective parameter freezing to the detailed algorithmic implementation, underlined by mathematical formulations for precision and clarity.

## 4. EXPERIMENTS

### 4.1 Experimental Setup

We conduct extensive experiments to evaluate the performance of LoRA-SP (Partial-Selective Low-Rank Adaptation), focusing on its efficacy in fine-tuning large language models (LLMs) while significantly reducing memory and computational overhead[19]. We compare LoRA-SP against baseline methods, including full-parameter fine-tuning (FT) and the original LoRA, across various NLP tasks, such as natural language understanding (NLU), machine translation (MT), and natural language generation (NLG)[5].

### 4.2 Datasets and Benchmarks

Our evaluation spans across multiple benchmarks, including RoBERTa models on GLUE benchmark tasks, T5 model for WMT16 English-Romanian translation, and LLaMA-2 for MMLU. Each dataset targets different aspects of model capabilities, providing a comprehensive assessment of LoRA-SP's versatility and efficiency[6].

### 4.3 Implementation Details

For LoRA-SP, we randomly select half of the parameters in matrices A and B for training, while the other half remains frozen, a technique we hypothesize to balance model flexibility and memory efficiency effectively. We use AdamW optimizer with a learning rate of 2e-5 and fine-tune for 3 epochs across all experiments. The rank size for LoRA-SP is set to r=16, chosen based on preliminary experiments indicating optimal trade-offs between performance and efficiency.

### 4.4 Result

**RoBERTa**

Our research delves into the fine-tuning performance of RoBERTa models on GLUE benchmark tasks, focusing on both the base model, with around 125 million parameters, and the large variant, housing approximately 355 million parameters. These models, which build upon BERT's architecture by optimizing fine-tuning processes and excluding the next-sentence prediction task, were tested for their natural language understanding (NLU) capabilities.

The experimentation began with a hyper-parameter optimization, primarily on the MRPC task, to establish ideal settings for subsequent tasks. The base model utilized a batch size of 64 and sequence length of 128, while the large model was fine-tuned with a batch size of 32 due to its larger size. These settings aimed to evaluate the models under uniform conditions.

Table 1 presents our findings, comparing the full fine-tuning (FT) approach with Low-Rank Adaptation (LoRA) and LoRA with Parameter Sharing (LoRA-SP). Noteworthy observations include the base model's FT achieving an average score of 83.8 across tasks. In contrast, LoRA and LoRA-SP methods, despite drastically fewer trainable parameters (0.9M and 0.45M, respectively), not only matched but occasionally surpassed FT performance, with LoRA-SP elevating the average to 84.5.

Table 1. This table presents a detailed analysis of the performance of RoBERTa models, both base and large variants, under various fine-tuning methods. Performance metrics include Matthews correlation for CoLA, Pearson correlation for STS-B, accuracy for SST-2, MRPC, QNLI, QQP, RTE, and both matched and mismatched MNLI scores, aiming for higher values as indicators of better model performance. Additionally, the table showcases the total number of trainable parameters for each configuration, highlighting the efficiency of different fine-tuning approaches like full training (FT) and low-rank adaptations (LoRA and LoRA-SP).

| Model&Method | #Trainable Parameters | MNKL | SST-2 | MRPC | CoLA | QNLI | QQP | RTE | STS-B |
|---|---|---|---|---|---|---|---|---|---|
| RoBBase(FT) | 121.3M | 86.5 | 92.1 | 91.3 | 58.9 | 91.2 | 90.9 | 69.8 | 89.9 |
| RoBBase(LoRA) | 0.9M | 86.7 | 93.4 | 90.6 | 62.4 | 92.3 | 88.9 | 70.2 | 90.3 |
| RoBBase(LoRA-SP) | 0.45M | 86.2 | 93.6 | 91.1 | 62.5 | 91.4 | 90.4 | 70.3 | 90.2 |
| RoBLarge(FT) | 348.M | 89.7 | 95.6 | 92.2 | 65.9 | 93.4 | 90.2 | 82.3 | 92.1 |
| RoBLarge(LoRA) | 1.8M | 90.3 | 96.5 | 91.2 | 65.6 | 93.6 | 91.3 | 82.1 | 91.3 |
| RoBLarge(LoRA-SP) | 0.9M | 90.4 | 96.4 | 92.3 | 66.2 | 93.7 | 91.3 | 82.5 | 92.3 |

For the RoBERTa-large variant, similar trends were observed. The FT method yielded an average score of 87.7, while the LoRA and LoRA-SP methods demonstrated that significant reductions in trainable parameters do not necessarily compromise, and can even enhance, performance across NLU tasks. Specifically, the LoRA-SP approach achieved an average performance score of 88.1 with just 0.9M trainable parameters, indicating efficiency and effectiveness in parameter utilization.

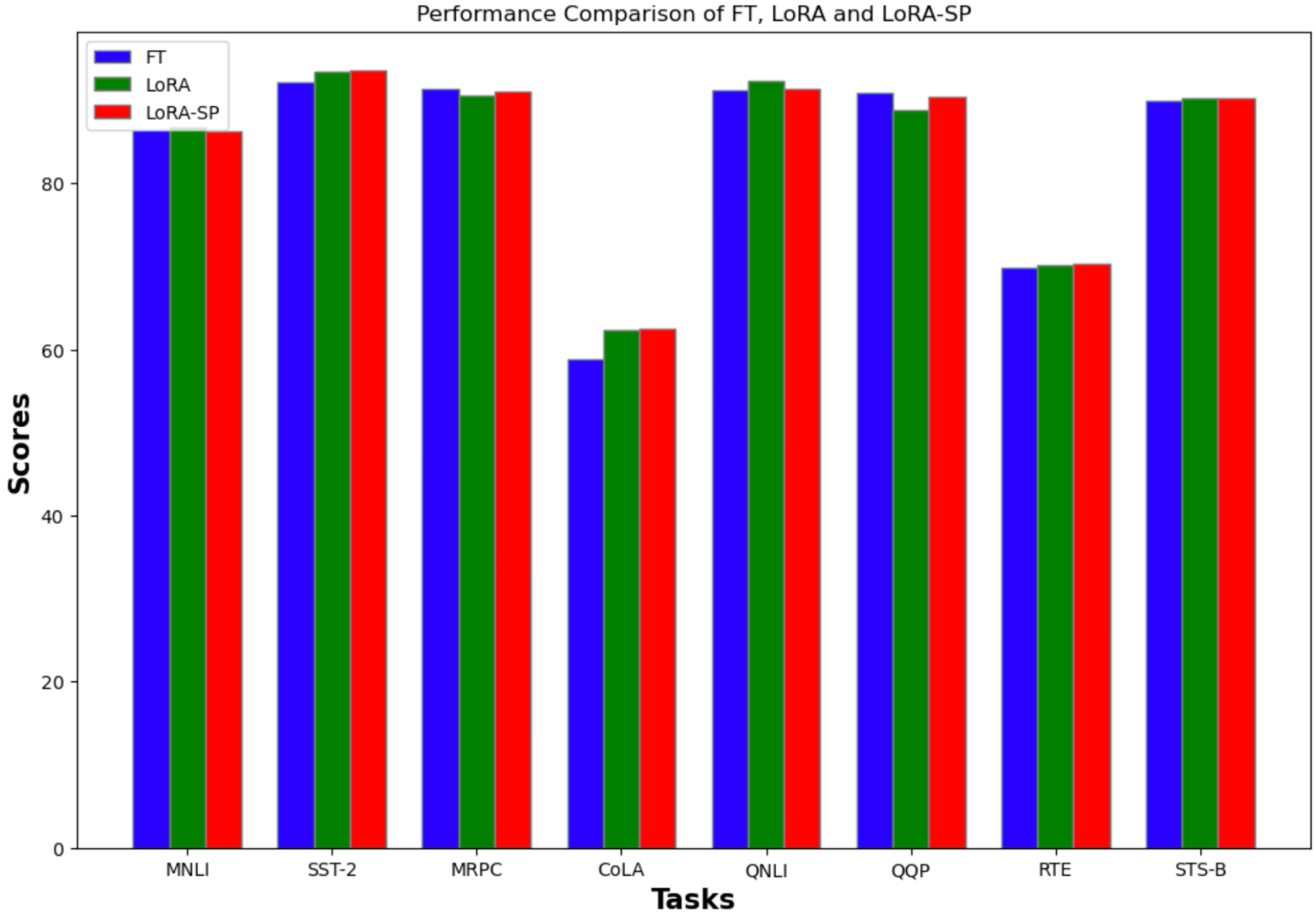

Figure 2. Performance Comparison of Full Fine-Tuning (FT), Low-Rank Adaptation (LoRA), and LoRA with Selective Parameter Freezing (LoRA-SP) across various NLP tasks. The figure illustrates that LoRA-SP achieves comparable performance to both FT and LoRA, despite requiring significantly fewer computational resources and lower memory occupation.

Our study underscores the efficacy of LoRA and LoRA-SP fine-tuning techniques in optimizing RoBERTa models for NLU tasks. These methods not only offer a pathway to maintain high performance with fewer parameters but also present a viable strategy for enhancing model efficiency and performance simultaneously. This balance between computational resourcefulness and task proficiency is critical, highlighting the potential for more sustainable model fine-tuning practices in the field of NLU[22].

**T5**

In the realm of machine translation (MT) and text generation, the T-5 (Text-to-Text Transfer Transformer) model stands as a pivotal development, offering a unified framework for addressing a wide array of NLP tasks through text-to-text processes. Our study extends to evaluating the fine-tuning performance of T-5 models across various sizes—small, base, and large—utilizing the WMT16 English-Romanian (En-Ro) dataset as a benchmark. This evaluation seeks to discern the effectiveness of different fine-tuning strategies, namely Full Fine-Tuning (FT), Low-Rank Adaptation (LoRA), and LoRA with Parameter Sharing (LoRA-SP), in enhancing translation quality as measured by BLEU and ROUGE scores.

Our analysis, encapsulated in Table 2, reveals insightful trends regarding the balance between model size, fine-tuning approach, and resultant performance. Notably, the T5-Base model fine-tuned with FT demonstrates a commendable BLEU score of 31.5 and ROUGE-L of 40.3, establishing a baseline for comparison. The adaptation through LoRA, while significantly reducing the model's trainable parameters to just 0.9M, slightly detracts from BLEU score performance yet marginally improves ROUGE-L to 40.6, suggesting a nuanced trade-off between model simplicity and translation fidelity. The LoRA-SP variant further refines this balance, achieving a BLEU score of 31.2 and a ROUGE-L of 39.8, indicating that even with a halved parameter count compared to LoRA, the model sustains its translation capabilities robustly[20].

Table 2. This table outlines the outcomes of fine-tuning T5 models of different sizes with various methodologies on the WMT16 English to Romanian (En-Ro) dataset. It provides a comparative view of the models' effectiveness by reporting BLEU scores and ROUGE metrics (ROUGE-1, ROUGE-2, and ROUGE-L), where higher scores reflect better translation quality. The table also indicates the number of trainable parameters for each model and method, offering insights into the models' complexity and the efficiency of the fine-tuning approaches.

| Model&Method | #Trainable Parameters | BLEU | ROUGE-1 | ROUGE-2 | ROUGE-L |
|---|---|---|---|---|---|
| T5Base(FT) | 218.8M | 31.5 | 42.6 | 28.9 | 40.3 |
| T5Base(LoRA) | 0.9M | 30.9 | 43.3 | 27.6 | 40.6 |
| T5Base(LoRA-SP) | 0.45M | 31.2 | 43.7 | 27.8 | 39.8 |
| T5Large(FT) | 725.4M | 33.4 | 51.2 | 33.2 | 47.3 |
| T5Large(LoRA) | 2.29M | 33.6 | 50.9 | 32.8 | 47.1 |
| T5Large(LoRA-SP) | 1.15M | 33.5 | 51.3 | 33.6 | 47.5 |

The T5-Large model's performance underlines the scalability of these fine-tuning methodologies. With FT, it reaches a BLEU score of 33.4 and a ROUGE-L of 47.3, showcasing the inherent capabilities of larger models in capturing linguistic nuances. However, the introduction of LoRA and LoRA-SP fine-tuning not only maintains but in some instances, slightly enhances performance metrics with a significantly reduced parameter footprint. For instance, LoRA-SP achieves a BLEU score of 33.5 and a ROUGE-L of 47.5, with just 1.15M trainable parameters, epitomizing the efficiency and efficacy of parameter-optimized fine-tuning.

These findings illuminate the potential of LoRA and LoRA-SP fine-tuning strategies in optimizing the performance of T-5 models across translation tasks. By dramatically reducing the number of trainable parameters without substantial losses in translation quality, these approaches underscore the possibility of achieving high linguistic performance in a more resource-efficient manner. This balance is crucial for advancing NLP technologies, especially in scenarios where computational resources are at a premium.

**LLaMA-2**

In our exploration, we fine-tuned LLaMA-7B and LLaMA-13B models using Full Fine-Tuning (FT), Low-Rank Adaptation (LoRA), and LoRA with Parameter Sharing (LoRA-SP) on the Alpaca dataset, focusing on 5-shot MMLU accuracy. The results, detailed in Table 3, demonstrate the efficacy of these strategies in enhancing model performance with a significant reduction in trainable parameters[17].

The LLaMA-7B model achieved a 5-shot MMLU accuracy of 39.8 with FT, while LoRA and LoRA-SP adaptations slightly reduced accuracy to 38.9 and 39.0, respectively, but with drastically fewer parameters (157.3M for LoRA and 78.7M for LoRA-SP). Similarly, the LLaMA-13B model fine-tuned with FT reached an accuracy of 47.8, with LoRA and LoRA-SP strategies achieving competitive accuracies of 45.3 and 45.4, respectively, further emphasizing the efficiency of these methods.

Table 3. In this table, we examine the impact of different fine-tuning strategies on the 5-shot MMLU accuracy of LLaMA models, specifically the 7B and 13B versions, using the Alpaca dataset. It provides a comparative analysis highlighting the base model performance and the improvements achieved through fine-tuning techniques (FT, LoRA, LoRA-SP), showcasing the effectiveness of each method. The table also details the total and trainable parameters, underscoring the scalability and adaptability of the LLaMA models in leveraging knowledge from limited examples for enhanced performance[16].

| Model&Method | #Trainable Parameters | #Total Parameters | 5-shot MMLU Accuracy |
|---|---|---|---|
| LLaMA-7B-Alpaca(FT) | 6624.4M | 6624.4M | 39.8 |
| LLaMA-7B-Alpaca (LoRA) | 157.3M | 6782.1M | 38.9 |
| LLaMA-7B-Alpaca (LoRA-SP) | 78.7M | 6702.1M | 39.0 |
| LLaMA-7B-Alpaca (FT) | 12986.8M | 12986.8M | 47.8 |
| LLaMA-13B-Alpaca (LoRA) | 248.6M | 13235.8M | 45.3 |
| LLaMA-13B-Alpaca (LoRA-SP) | 124.3M | 13112.3M | 45.4 |

This analysis underscores the potential of LoRA and LoRA-SP in maintaining high performance levels with reduced parameter counts, highlighting a trade-off between computational efficiency and accuracy. These findings contribute to the broader discussion on optimizing large language models for efficient adaptation to new tasks with limited data.

### 4.5 Discussion

Our experiments demonstrate that LoRA-SP not only achieves competitive performance across a range of NLP tasks but also significantly enhances memory and computational efficiency. By selectively training half of the parameters in the low-rank matrices A and B, LoRA-SP presents a promising approach for fine-tuning LLMs, especially in resource-constrained environments[9].

### 4.6 Analysis of Selective Parameter Training

A deeper analysis into the selective training of parameters in LoRA-SP reveals that freezing half of the parameters in A and B does not impede the model's ability to adapt to new tasks. Instead, it directs the model's learning capacity towards the most beneficial updates, thereby optimizing memory and computational resources without sacrificing performance.

## 5. CONCLUSION

In this study, we introduced LoRA-SP, a novel approach that selectively freezes half of the parameters in the A and B matrices during the fine-tuning of large language models (LLMs) using Low-Rank Adaptation (LoRA). Our experimental results demonstrate that LoRA-SP not only retains the high performance of traditional full-parameter fine-tuning methods but also significantly reduces the computational overhead and memory requirements. This efficiency gain opens new avenues for deploying LLMs in resource-constrained environments, making state-of-the-art NLP capabilities more accessible.

Moreover, LoRA-SP's selective parameter freezing strategy contributes to a deeper understanding of parameter efficiency in LLMs. It highlights the potential of partial parameter updates as a viable path toward more sustainable and efficient model fine-tuning practices. As the field of NLP continues to evolve, the implications of our findings suggest that further exploration into selective and adaptive parameter updates could yield even more efficient and effective fine-tuning methodologies.

Future work will explore extending LoRA-SP to a broader range of LLM architectures and fine-tuning scenarios. Additionally, investigating the impact of different selection strategies for freezing parameters and their correlation with specific model tasks and domains could further optimize the efficiency and performance of LLM fine-tuning. Our research sets the stage for more sophisticated approaches to model adaptation, emphasizing the balance between performance, efficiency, and resource utilization.

## REFERENCES

[1] Brown, T., Mann, B., Ryder, N., Subbiah, M., Kaplan, J. D., Dhariwal, P., Neelakantan, A., Shyam, P., Sastry, G., Askell, A., Agarwal, S., Herbert-Voss, A., Krueger, G., Henighan, T., Child, R., Ramesh, A., Ziegler, D., W u, J., Winter, C., … Amodei, D. (2020). Language models are few-shot learners. Advances in Neural Informatio n Processing Systems, 33, 1877–1901. https://proceedings.neurips.cc/paper_files/paper/2020/hash/1457c0d6bfc b4967418bfb8ac142f64a-Abstract.html?utm_medium=email&utm_source=transaction

[2] Chen, J., Zhang, A., Shi, X., Li, M., Smola, A., & Yang, D. (2023). Parameter-efficient fine-tuning design space s (arXiv:2301.01821). arXiv. https://doi.org/10.48550/arXiv.2301.01821

[3] Chen, Y., Qian, S., Tang, H., Lai, X., Liu, Z., Han, S., & Jia, J. (2023). Longlora: Efficient fine-tuning of long-c ontext large language models (arXiv:2309.12307). arXiv. https://doi.org/10.48550/arXiv.2309.12307

[4] Ding, N., Qin, Y., Yang, G., Wei, F., Yang, Z., Su, Y., Hu, S., Chen, Y., Chan, C.-M., Chen, W., Yi, J., Zhao, W., Wang, X., Liu, Z., Zheng, H.-T., Chen, J., Liu, Y., Tang, J., Li, J., & Sun, M. (2023). Parameter-efficient fin e-tuning of large-scale pre-trained language models. Nature Machine Intelligence, 5(3), 220–235. https://doi.org /10.1038/s42256-023-00626-4

[5] Fu, Z., Yang, H., So, A. M.-C., Lam, W., Bing, L., & Collier, N. (2023a). On the effectiveness of parameter-effi cient fine-tuning. Proceedings of the AAAI Conference on Artificial Intelligence, 37(11), 12799–12807. https:// doi.org/10.1609/aaai.v37i11.26505

[6] Zhang, Y., Gong, Y., Cui, D., Li, X., & Shen, X. (2024). Deepgi: An automated approach for gastrointestinal tra ct segmentation in mri scans (arXiv:2401.15354). arXiv. http://arxiv.org/abs/2401.15354

[7] Houlsby, N., Giurgiu, A., Jastrzebski, S., Morrone, B., Laroussilhe, Q. D., Gesmundo, A., Attariyan, M., & Gell y, S. (2019). Parameter-efficient transfer learning for nlp. Proceedings of the 36th International Conference on Machine Learning, 2790–2799. https://proceedings.mlr.press/v97/houlsby19a.html

[8] Hu, E. J., Shen, Y., Wallis, P., Allen-Zhu, Z., Li, Y., Wang, S., Wang, L., & Chen, W. (2021). Lora: Low-rank a daptation of large language models (arXiv:2106.09685). arXiv. http://arxiv.org/abs/2106.09685

[9] Huang, X., Zhang, Z., Guo, F., Wang, X., Chi, K., & Wu, K. (2024). Research on older adults' interaction with e-health interface based on explainable artificial intelligence (arXiv:2402.07915). arXiv. http://arxiv.org/abs/240 2.07915

[10] Lermen, S., Rogers-Smith, C., & Ladish, J. (2023). Lora fine-tuning efficiently undoes safety training in llama 2 -chat 70b (arXiv:2310.20624). arXiv. https://doi.org/10.48550/arXiv.2310.20624

[11] Li, Y., Yu, Y., Liang, C., He, P., Karampatziakis, N., Chen, W., & Zhao, T. (2023). Loftq: Lora-fine-tuning-awa re quantization for large language models (arXiv:2310.08659). arXiv. https://doi.org/10.48550/arXiv.2310.0865 9

[12] Liu, Q., Wu, X., Zhao, X., Zhu, Y., Xu, D., Tian, F., & Zheng, Y. (2023). Moelora: An moe-based parameter eff icient fine-tuning method for multi-task medical applications (arXiv:2310.18339). arXiv. https://doi.org/10.4855 0/arXiv.2310.18339

[13] Liu, Y., Ott, M., Goyal, N., Du, J., Joshi, M., Chen, D., Levy, O., Lewis, M., Zettlemoyer, L., & Stoyanov, V. (2 019). Roberta: A robustly optimized bert pretraining approach (arXiv:1907.11692). arXiv. https://doi.org/10.485 50/arXiv.1907.11692

[14] Ouyang, L., Wu, J., Jiang, X., Almeida, D., Wainwright, C., Mishkin, P., Zhang, C., Agarwal, S., Slama, K., Ra y, A., Schulman, J., Hilton, J., Kelton, F., Miller, L., Simens, M., Askell, A., Welinder, P., Christiano, P. F., Lei ke, J., & Lowe, R. (2022). Training language models to follow instructions with human feedback. Advances in Neural Information Processing Systems, 35, 27730–27744. https://proceedings.neurips.cc/paper_files/paper/202 2/hash/b1efde53be364a73914f58805a001731-Abstract-Conference.html

[15] Wang, G., Gong, Y., Zhu, M., Yuan, J., & Wei, K. (2023). Unveiling the future navigating next-generation ai fr ontiers and innovations in application. International Journal of Computer Science and Information Technology, 1(1), 147–156. https://doi.org/10.62051/ijcsit.v1n1.20

[16] Wang, X., Aitchison, L., & Rudolph, M. (2023). LoRA ensembles for large language model fine-tuning (arXiv: 2310.00035). arXiv. https://doi.org/10.48550/arXiv.2310.00035

[17] Wang, Y., Xu, L., Liu, W., Li, R., & Gu, J. (2023). Network intrusion detection based on explainable artificial i ntelligence. Wireless Personal Communications, 131(2), 1115–1130. https://doi.org/10.1007/s11277-023-10472 -7

[18] Wu, K., & Chi, K. (2024). Enhanced e-commerce customer engagement: A comprehensive three-tiered recomm endation system. Journal of Knowledge Learning and Science Technology ISSN: 2959-6386 (Online), 2(3), 348 –359. https://doi.org/10.60087/jklst.vol2.n2.p359

[19] Yang, Y., UY, M. C. S., & Huang, A. (2020). Finbert: A pretrained language model for financial communicatio ns (arXiv:2006.08097). arXiv. https://doi.org/10.48550/arXiv.2006.08097

[20] Zhang, L., Zhang, L., Shi, S., Chu, X., & Li, B. (2023). Lora-fa: Memory-efficient low-rank adaptation for large language models fine-tuning (arXiv:2308.03303). arXiv. https://doi.org/10.48550/arXiv.2308.03303

[21] Zhang, Y., Zhu, M., Gong, Y., & Ding, R. (2023). Optimizing science question ranking through model and retri eval-augmented generation. International Journal of Computer Science and Information Technology, 1(1), 124–130. https://doi.org/10.62051/ijcsit.v1n1.17

[22] Zhou, H., Lou, Y., Xiong, J., Wang, Y., & Liu, Y. (2023). Improvement of deep learning model for gastrointesti nal tract segmentation surgery. Frontiers in Computing and Intelligent Systems, 6(1), 103–106. https://doi.org/1 0.54097/fcis.v6i1.19

[23] Srivastava, N., Hinton, G., Krizhevsky, A., Sutskever, I., & Salakhutdinov, R. (2014). Dropout: a simple way to prevent neural networks from overfitting. *The journal of machine learning research*, *15*(1), 1929-1958. https://w ww.jmlr.org/papers/volume15/srivastava14a/srivastava14a.pdf?utm_content=buffer79b43&utm_medium=socia l&utm_source=twitter.com&utm_campaign=buffer,